# Optimization of IoT-Enabled Physical Location Monitoring Using DT and VAR

Ajitkumar Sureshrao Shitole, Amity University, Mumbai, India

https://orcid.org/0000-0002-0310-277X

Manoj Himmatrao Devare, Amity University, Mumbai, India

**ABSTRACT**

This study shows an enhancement of IoT that gets sensor data and performs real-time face recognition to screen physical areas to find strange situations and send an alarm mail to the client to make remedial moves to avoid any potential misfortune in the environment. Sensor data is pushed onto the local system and GoDaddy Cloud whenever the camera detects a person to optimize the physical location monitoring system by reducing the bandwidth requirement and storage cost onto the cloud using edge computation. The study reveals that decision tree (DT) and random forest give reasonably similar macro average f1-scores to predict a person using sensor data. Experimental results show that DT is the most reliable predictive model for the cloud datasets of three different physical locations to predict a person using timestamp with an accuracy of 83.99%, 88.92%, and 80.97%. This study also explains multivariate time series prediction using vector auto regression that gives reasonably good root mean squared error to predict temperature, humidity, light-dependent resistor, and gas time series.

## 1. INTRODUCTION

Internet of Things (IoT) is one of the promising and quickly rising advances in the field of Information Technology and Communication Engineering. Bunches of gadgets can be associated with the assistance of IoT to impart and trade their data and information. In the present life, it is important to screen the physical area with the assistance of IoT, where quantities of various sensors are associated with Single Board Computer (SBC). Investigation of the physical area is required to recognize any irregular conditions in situations like home areas, open territories like railroad stations, air terminals, clinics, instructive establishments, delicate research facilities, businesses, and so on. Unusual conditions can be abrupt diminish or increment in temperature and dampness, uncommon varieties in the force of light, the unexpected increment in gas sensor esteem, obscure individual's identification in the premises, which thus can make extreme harm to the area and environment. A sudden increase in the intensity of the light may be because of natural or intentional or unintentional light exposure, but the values of the Passive Infra-Red (PIR) sensor or face recognition can identify the abnormal condition in the environment. In pharmacy industries, certain drugs and medicines are placed away from the high exposure of the light intensity, which may cause a reaction to medicines and leads to damage to the same. So the analysis of such sensor values is required to avoid the loss.

The PIR sensor detects the movement of the object around the system. This is necessary to know the presence, proximity, and occupancy of the person in the environment, which may become the confirmatory test for unauthorized access in the surroundings. Drastic variations in the temperature or humidity may cause damage to the persons who are suffering from cold or hot related disorders like respiratory infections, eye infections, and muscular spasms, and so on. IoT empowered framework with mixed media information, for example, advanced pictures of human faces are helpful for face discovery and acknowledgment. Face acknowledgment is valuable in different situations, for example, interruption recognition, distinguishing a few activities, for example, switching ON/OFF different gadgets, recognizing client's propensities in the earth to be familiar with when the shopper is at the house, and connecting with the gadgets, etc. The improvement of IoT empowered frameworks with face acknowledgment rolls out a noteworthy improvement in the wellbeing and security of premises. The more strong and ground-breaking structure can be accomplished with the assistance of IoT and face acknowledgment.

The proposed structure uses the sensor to request a sentiment of the earth. IoT, with the support of Machine Learning (ML), is used to alert the position when the individual is in certifiable peril. It uses Raspberry-Pi as the basic base of undertaking for planning data. In this work, the design and analysis of IoT enabled Physical Location Monitoring System (PLMS) is carried out to collect real-time sensor data along with person detection and recognition to perform person prediction using sensor data and timestamp with ML algorithms. This paper is focused on various research questions like detection of abnormal conditions in the physical location, how the sensor data is useful to predict a person, finding the most useful sensors to predict a person, finding the most effective ML technique to predict a person using a timestamp, and prediction of sensor values of different time series using Vector Auto Regression (VAR) model in a multivariate scenario. Three different physical locations are considered here for collecting the information from the sensors to extract knowledge or useful information from it. The remaining part of this paper is organized as follows: Section 2 discusses the related work in the field of IoT, ML, and sensor data analysis using ML, model selection, and evaluation. Section 3 explains the proposed architecture, which tells how the PLMS has been implemented. Section 4 gives the insights of data analysis in terms of person prediction using sensor data, feature importance, person prediction using a timestamp, and prediction of sensor values of different time series in a multivariate scenario. Finally, the paper is concluded in section 5 with limitations and the future scope of the PLMS.

## 2. RELATED WORK

This section discusses research work to monitor the physical location using IoT, sensor data analysis using ML, imbalanced class problems for binary and multi-class classification, time series prediction and analysis, and model selection and evaluation techniques.

Rathod et al. (2017) developed a real-time home monitoring system to control various appliances of the home which is useful not only for normal users but also for handicapped people who are close or remote to the system and make their life comfortable. Raspberry-Pi is used to control fans and lights, but the system is deficient in terms of sending an alert mail or message, multimedia data, edge computation, multithreading, pushing data on to the Cloud for further processing, data analysis to extract meaningful information from the sensor data. Joshi et al. (2017) monitored physical location using Raspberry-Pi and Arduino boards to control various appliances using sensors. Readings were also sent to the ThingSpeak IoT open source Cloud to monitor the location remotely and an alert mail was delivered to the user if a threat was found. Although the location was monitored using the sensors, the system does not involve the use of edge computation, multithreading, pushing data onto the Cloud, data analysis using ML. Mandula et al. (2015) monitored physical location in real-time using an Arduino board to control fans and lights. The home was controlled in an indoor environment using Bluetooth, while Ethernet was used in an outdoor environment. The authors did not address the involvement of alert mail or message, edge computation, multithreading, pushing data on to the Cloud for further processing, data analysis to extract meaningful information from the sensor data. Kodali et al. (2016) introduced IoT based smart security to home automation using a PIR sensor and microcontroller. The system was dependent on the user's discretion and judges the ability of the situation whether the person is a guest or intruder. Vinay Sagar et al. (2015) introduced the Home Automation System (HAS) utilizing Intel Galileo that utilizes the reconciliation of cloud organizing, remote correspondence, to give the client remote control of different lights, fans, and apparatuses inside their home and putting away the information in the cloud. The framework will consequently change based on sensors' information. Although the authors used various sensors such as temperature, gas, motion, and light-dependent resistor to control the home appliances automatically, there was no involvement of multimedia data, multithreading, edge commutation, and data analysis. Folea et al. (2012) used the Tag4M device to create a smart home by monitoring temperature, humidity, light sensors, which were applied to control fans and light automatically with the help of threshold values of the sensors. Moraru et al. (2010) developed sensor-based techniques to predict the number of people in the environment using sensor data aligned with manual data. Sensor data consisted of temperature, humidity, pressure, and light intensity, while manual data consisted of the number of persons present, the number of computers running, and whether the window is closed or opened. The developed system by Moraru et al. (2010) does not discuss a prediction of a person's name using sensor data and timestamp. It also does not discuss a prediction of sensor values using ML. Devare (2018) investigated that a huge amount of data captured from the sensor needs to be processed to get an insight into the data. The author also discussed synchronization and handshaking problems of the sensors and libraries installed into the board that can be analyzed through the numerical tools and techniques. The system neither addressed the involvement of multithreading to synchronize sensor data with multimedia data nor the prediction of person and sensor values using a timestamp.

Rao and Uma (2015) proposed a Raspberry-Pi based home monitoring system to control the home appliances using mobile, which helps the senior citizens as well as handicapped persons to make their life comfortable. The security system was also designed using webcam surveillance for intrusion detection to take necessary action. The developed system has a scarcity of edge computation and collection of sensor data either into the local or the Cloud system to perform further data analysis.

The Research gap and contribution of the IoT enabled physical location monitoring is summarized in Table 1. The experimentation and analysis studied in this paper mainly focused on resolving all the research gaps, which are specified in Table 1. This paper solves the research gaps by utilizing edge computation to optimize storage cost onto the Cloud and multi-threading approach to synchronize sensor data with multimedia data to perform real-time face recognition as well as by pushing sensor data onto the Cloud to perform data analysis to optimize person prediction using sensor data and timestamp.

Researchers monitored the home physical location to identify an abnormal condition in the environment to take corrective action to avoid future loss in the premises and optimized person recognition to determine whether the recognized person is valid or invalid with Decision Tree (DT) accuracy of 89.91% using stratified 10 fold cross-validation (Shitole and Devare, 2019). Heyns et al. (2019) demonstrated that traffic stages can be ordered utilizing driving conduct. Drivers conduct changes as the traffic stage changes and these progressions can correspond to these traffic stages utilizing machine learning with an accuracy of 95% and precision of 92%. Kaivonen and Ngai (2019) presented an experimental approach towards real-time monitoring of air pollution on community transport vehicles which uses IoT to measure pollution level. Experiments were conducted to measure data and communication quality of the system. Tziortzioti and Amaxilatis (2019) proposed a new approach using IoT sensors for estimating different ocean water parameters, are investigated here, pointing towards an instructive setting, to prompt a more profound comprehension of the utilization of amphibian conditions as regular resources, and towards the appropriation of ecologically well-disposed practices. Kakhki et al. (2019) proposed new models which can foresee damage seriousness order dependent on harmed body part, body gathering, and nature of damage, nature gathering, the reason for the damage, cause gathering, and age and residency of harmed laborers with the precision rate of 92–98%. Wrzesień et al. (2019) investigated the control of apple scab using machine learning algorithms with the help of meteorological parameters like temperature, wind speed, and relative humidity. Sensor data was collected for two rising seasons from wetness sensors which were planted in four parts in the tree canopy.

**Table 1. Research gap and contribution of the IoT enabled physical location monitoring**

| Author & Paper Title | Research Gap | Contribution of the PLMS |
|---|---|---|
| Rathod et al. (2017), "Raspberry Pi Based Home Automation Using WiFi, IoT, and Android for Live Monitoring". | Lack of multimedia data, edge computation, multithreading, pushing data on to Cloud for further processing, and data analysis to extract meaningful information from the sensor data. | Incorporated sensor data with multimedia data to create more robust IoT enabled PLMS to perform further data analysis. |
| Joshi et al. (2017), "Performance Measurement & IoT Monitoring for Smart Home". | Deficient in terms of edge computation, multithreading, pushing data onto Cloud, and data analysis using ML. | ● Created effective and efficient IoT enabled optimized PLMS to identify abnormal conditions in the environment using ML supported statistical tools and threshold values of the sensors.<br>● Built the most consistent and reliable DT ML model to predict a person's name using sensor data and identified the most contributed sensor to predict the same.<br>● Constructed the most efficient predictive DT ML model to predict a person using a timestamp. |
| Kodali et al. (2016), "IoT Based Smart Security and Home Automation System". | Short of multimedia data, edge computation, multithreading, pushing data onto Cloud, and data analysis using ML. | |
| Mandula et al. (2015), "Mobile based home Automation using Internet of Things (IoT)". | Absence of an alert mail, sensors, multimedia data, edge computation, multithreading, and pushing data on to Cloud for further processing. | |
| Vinay Sagar et al. (2015), "Home Automation Using Internet of Things". | Scarcity of multimedia data, edge computation, multithreading, and sensor data analysis. | |
| Folea et al. (2012), "Smart Home Automation System Using Wi-Fi Low Power Devices". | Absence of an alert mail, multimedia data, edge computation, multithreading, pushing data on to Cloud for further processing, and sensor data analysis. | |
| Moraru et al. (2010), "Using ML on Sensor Data". | Nonexistence of multimedia data, edge computation, and multithreading, pushing data onto Cloud. | ● Pushed sensor data on to the Cloud and performed data analysis to predict a person using sensor data and timestamp.<br>● Constructed the VAR ML model to predict sensor values of the different time series such as temperature, humidity, gas, and LDR in a multivariate scenario. |
| Devare (2018), "Analysis and Design of IoT Based Physical Location Monitoring System". | No involvement in multimedia data in terms of the face image, edge computation, multithreading, and sensor data analysis. | Used a multithreading approach to solve sensor synchronization problems with face recognition and performed sensor data analysis. |

Muller and Guido (2017a) presented how to apply python using scikit-learn to apply different concepts of ML-like supervised learning, unsupervised learning, and representation of data and engineering features, model evaluation and improvement, time series analysis and so on to complete the research activity effectively. Raschka (2015a) also explained the practical approach of data mining basic concepts, ML algorithms, and evaluation of the models and so on using python programming. Hossin and Sulaiman (2015) presented that not only accuracy is the performance matrix of the classifier but also other metrics are available and need to apply as per the type of the dataset to improve the performance of the models to get proper results from the datasets. Gu et al. (2009) represented that an imbalanced dataset may lead to misinterpretation of the classification results and tend to wrong decisions. Researchers explained other performance measures apart from the accuracy to handle class imbalance problems. Joshi (2002) explained the importance of rare class and how to predict rare class effectively using machine learning. Experiments showed that performance metric recall and precision gave good predictive results. Burnap et al. (2017) introduced a new technique to perform a multi-class classification of suicide-related data available on twitter to classify into 7 different classes. The experiment showed that f-measure to represent the overall performance of the system was 0.728 and 0.69 for the suicidal ideation class. Tsoumakas and Katakis (2009) proposed the work of multi-class classification and organized diverse literature in proper format and evaluated certain multi-class classification methods in the form of accuracy, recall, and precision. Tharwat (2018) introduced various binary and multi-classification performance measures in the form of the confusion matrix, accuracy, recall, precision, ROC, sensitivity, specificity for balanced as well as imbalanced datasets. Ghanem et al. (2010) focused on two major challenges in ML-like multi-class classification and imbalanced class problems. Authors applied One-Against-One and One-Against-All approaches to overcome the difficulties of both problems and got reasonable results of the multi-class classifiers for pattern recognition. Sahare and Gupta (2012) outlined a review of the multi-class classification of imbalanced datasets. Researchers discussed the problems of prediction accuracy and voting of correct class to unknown samples. Approaches like extending the binary tree to decompose the multi-class problem are applied. Other techniques like one-vs-one and one-vs-all are also applied to solve the problem of imbalanced data for multi-class classification problems. Jeatrakul and Wong (2012) presented that the solutions of the class imbalance problem of binary classifiers cannot be directly applied to solve the issues of the class imbalance problem of multi-class classification problems and researchers combined the One-Against-All technique with data balancing to increase the overall performance of the classifiers.

Brownlee (2019) explained how to prepare the data to create the various models to predict the future using time series as an input component in python. The author explained different basic concepts of time series forecasting like time series, time series nomenclature, components of time series, and examples of time series. Ostashchuk (2017) presented a master's thesis which includes various concepts of time series prediction and analysis. The thesis shows the necessary steps of data preprocessing and different forecasting models to perform data analysis to predict the results based on the time component. Kohavi (1995) explained various model accuracies estimation methods like holdout, random subsampling, bootstrap, cross-validation, k-fold cross-validation, leave-one-out cross-validation, and stratified k-fold cross-validation. Experimental results showed that among various model selection methods in ML, stratified 10-fold cross-validation gives better results for the classifiers, and the same is recommended both in terms of variance and bias. Han et al. (2012a) presented in their book various model selection methods and it has been suggested that to select models effectively, stratified 10-fold cross-validation is recommended as it distributes data samples of different labels evenly into different folds to increase the performance of the models. Raschka (2015b) also explained various model evaluation and selection methods with examples using python programming. For imbalanced class datasets stratified 10 fold cross-validation gives very good performance because of its property of even distribution of imbalanced class samples into different folds to get low bias and variance. Muller and Guido (2017b) also discussed model evaluation and improvement methods to select the model which is more reliable. It has been specified that stratified 5-fold or 10-fold cross-validation splits the imbalanced class data in such a way that each fold gets the same proportion between the imbalanced classes to achieve less variance and bias to get a more reliable model.

## 3 PROPOSED ARCHITECTURE

Physical area conditions are checked by various sensors like Digital Humidity Temperature (DHT) sensor (for observing temperature and moistness individually), Light Dependent Resistor (LDR) (for identifying power of light), Gas sensor (for recognizing spillage of gas), and PIR (for distinguishing movement of an object). The camera snaps the picture of an individual when PIR identifies movement to distinguish any unapproved access in the environment. This section consists of four sub-sections. The first sub-section discusses the block diagram of the IoT enabled PLMS. The second sub-section illustrates face recognition, creation of multiple threads to read different sensor values, and to push real-time sensor data along with recognized person names onto the Cloud for further data analysis. The real-time sensor dataset is described in the third sub-section followed by data preprocessing and feature scaling in the last sub-section.

### 3.1 Block Diagram of the IoT Enabled PLMS

Figure 1 shows the block diagram of the IoT enabled PLMS in which few sensors are associated with Raspberry Pi through Analog to Digital Converter (ADC), and others are straightforwardly associated with it. Raspberry Pi is utilized to bring information from various sensors and transfer onto the ThingSpeak server, GoDaddy Cloud, and on the nearby framework. To accomplish synchronization between various sensors and web camera readings, multithreading is consolidated utilizing Python programming language with the assistance of the anaconda Jupyter environment. Edge computation is performed in the Raspberry-Pi itself such that whenever the camera detects and recognizes a person, only name of the recognized person in textual form is aligned with sensor data and pushed onto the Cloud to optimize the storage bandwidth requirement and storage cost onto the Cloud to create the Cloud dataset. All sensor readings, regardless of individual identification, are additionally recorded into the nearby server with a particular sampling rate to create a local dataset. As shown in Figure 1, data analysis is also performed over Cloud as well as local datasets to extract useful information from it using different supervised ML algorithms. Any IoT enabled PLMS can use the output of the data analysis, i.e., optimized sensor data analysis technique, to predict a person using sensor data and timestamp separately. The readings are additionally put away onto the ThingSpeak server to plot the sensor esteems as live charts to screen the physical area adequately. ThingSpeak is the IoT open-source stage where the sensor information can be gathered, prepared, broken down, and pictured utilizing HyperText Transfer Protocol (HTTP) with the assistance of the Internet or Local Area Network (LAN). The web page is also created to display all sensor values along with the name of the person detected to indicate the status of the environment, either normal or abnormal. If an irregular condition has happened in the environment, an alarm mail is additionally sent to the client to make a restorative move to stay away from further misfortune in the environment.

Figure 1. Block diagram of the IoT enabled PLMS

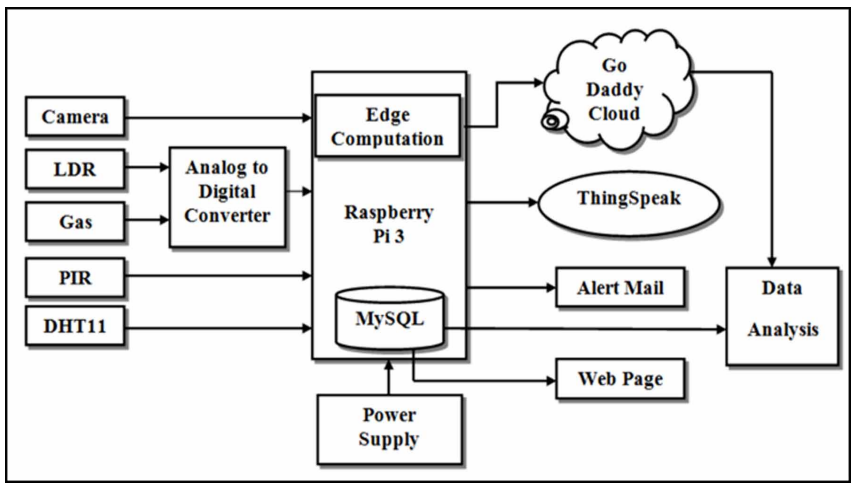

### 3.2 The IoT Enabled PLMS with Face Recognition

To screen physical areas, the sensor information is caught routinely in a constant manner and put away onto the nearby server. At whatever point a human face is distinguished and perceived as either a known or obscure individual, similar information is put away onto GoDaddy Cloud Service for further use and examination. Individual acknowledgment and sensor readings are caught simultaneously utilizing multithreading programming in python. To catch sensor esteems and to perceive the human face progressively, multithreading encoding in Python is applied as Raspberry Pi 3 has a quad-core processor. Two additional threads, along with the main thread, are formed to get concurrent processing, which in turn, to get utmost throughput. The main thread, i.e., Algorithm 1, is used to capture the live image frame by frame and to perform processing on that captured image for face recognition. The main thread is also used to push sensor data along with a timestamp and recognized face-name onto GoDaddy Cloud for further processing. The first thread, out of two additional threads, is used to read humidity and temperature sensor values. The second thread is used not only to read Gas, LDR, and PIR sensor values but also used to push all sensor values onto the local system to create a local database and onto ThingSpeak IoT platform to plot sensor values in real-time. Certain threshold values are set to various sensors to identify abnormal conditions in the location, and if it has occurred, an alert mail is sent to the user using the second thread to take necessary action to avoid any future loss in the environment. A face recognition library that recognizes and manipulates faces from Python is installed onto the system. The local database of known faces is created to compare with live captured images frame by frame.

Algorithm 1 shows procedural steps of main thread for face recognition, creation of two threads to read different sensor values and push data onto the Cloud. First, necessary libraries such as cv2, smtplib, face_recognition, threading, and so on are imported. Step 3 creates the object of the VideoCapture to capture a video with parameter 0 as one external camera is connected to the PLMS. Known person's images are loaded into respective variables from the database, and their face encodings are stored into respective variables in steps 4 and 5 respectively. Two arrays are initialized with known person face encodings and known person names for further use in steps 13 and 14. Two threads viz. t1 and t2 are created, and their execution is started with a delay of 4 seconds in steps 6 and 8 respectively. Capturing of frames and processing of frames is completed in steps 9 and 10 respectively, in which frame is captured and converted to a small Red Green Blue (RGB) form to process it. Face locations and encodings of captured images are recorded into variables in steps 11 and 12 respectively. Captured frames' face encodings are compared with known face encodings to find the name of the person detected in step 13. If the match is found, the person variable is set to known person name otherwise it is set to 'Unknown' in step 14. The face with the name of the person detected is displayed on the screen with a box around it in step 15. Detected person name with sensor data is also pushed onto GoDaddy Cloud with a specific sampling rate in step 16 to create Cloud dataset for further processing. The procedure is repeated in step 17 to continuously monitor the physical location in real-time.

**Algorithm 1** Face recognition and multiple threads
1. **Start**
2. Import necessary libraries such as cv2, smtplib, face_recognition, threading, etc.
3. Create an object of VideoCapture ().
4. Load the images of known persons.
5. Find encodings of the known images using the built-in function of the face_recognition library.
6. Create the first thread to read temperature and humidity sensor values using DHT11 sensor
7. Set a delay of 4 seconds.
8. Create a second thread to read LDR, Gas, and PIR sensor values.
9. Capture the frame of video using an externally connected camera.
10. Resize the frame and convert from Blue Green Red (BGR) to RGB color.
11. Find the locations of newly captured images.
12. Find the face encodings of the newly captured image using locations found in step 11.
13. Compare face encodings of captured new images with the known face encodings.
14. If the match is found store name of the recognized person otherwise store 'Unknown' label.
15. Resize the frame and display the name of the recognized person around the face.
16. Push sensor data along with the name of the recognized person onto GoDaddy Cloud.
17. Repeat the procedure with a delay of 4 seconds to get synchronization between sensor readings using additional threads and face recognition.
18. **End**

### 3.3 Portion of Real-time Collected Sensor Dataset of the PLMS

The prime focus of the PLMS is to detect abnormal conditions in the environment using different sensors and get to know the most useful sensors to classify or predict a person. Usually, abnormal conditions occur in the environment due to drastic variations in temperature, humidity, the intensity of light, gas leakage, and unauthorized access to the premises. To achieve proper synchronization between different sensor readings and face recognition, a multi-threading programming approach with a specific sampling rate is used. The PLMS may get little bit variations in sensor readings due to sensors' sampling rate, response time, and synchronization between sensors and face recognition at the same data and time. Table 2 shows the portion of the IoT enabled physical location monitoring sensor datasets. Dataset is labeled and consists of seven features such as Person, Temp ($^0$C), LDR (Lux), Gas (PPM), PIR, Hum (%), and Timestamp. Temp, LDR, Gas, and Hum are numeric features, whereas Person and PIR are categorical features. Timestamp feature is available in the form of date and time. All numeric features are having a different scale of measurements. Temperature is measured in Degree Celsius ($^0$C); Hum is measured in percentage (%); LDR is measured in Lux, and Gas is measured in Parts per Million (PPM). Person categorical feature has five different categories like 'Ajitkumar', 'Yogita', 'Pramila', 'Swaroop', and 'Unknown'. All these categories of Person features are nothing but the names of the persons detected in the home physical location. PIR feature has two categories like 'Yes' and 'No'. 'Yes' category indicates motion in the location is detected and the 'No' category means motion is not detected. After removal of Timestamp features, the dataset is used for multi-class classification with Person feature as an output feature and Temp, LDR, Gas, Hum, and PIR as an input features to perform optimization of person prediction using sensor data analysis and observe the performance of the classifiers to optimize the PLMS.

**Table 2. A portion of IoT enabled physical location monitoring sensor dataset**

| Person | Temp (°C) | LDR (Lux) | Gas (PPM) | PIR | Hum (%) | Timestamp |
|---|---|---|---|---|---|---|
| Ajitkumar | 26 | 299.4 | 0.11 | No | 66.86 | 2018-09-13 10:59:19.319301 |
| Ajitkumar | 26 | 277.2 | 0.1 | Yes | 66 | 2018-09-13 10:59:25.436754 |
| Ajitkumar | 26 | 296.3 | 0.1 | No | 67 | 2018-09-13 10:59:31.036016 |
| Ajitkumar | 26 | 291.05 | 0.1 | Yes | 67 | 2018-09-13 10:59:41.162160 |
| Ajitkumar | 26 | 280 | 0.1 | No | 67 | 2018-09-13 10:59:46.225767 |
| Ajitkumar | 26 | 273.2 | 0.1 | Yes | 67 | 2018-09-13 10:59:51.290994 |
| Swaroop | 26 | 266.57 | 0.11 | Yes | 66.91 | 2018-09-13 11:03:06.093508 |
| Swaroop | 26 | 261.15 | 0.12 | No | 66.75 | 2018-09-13 11:03:24.441326 |
| Swaroop | 26 | 260 | 0.13 | Yes | 67 | 2018-09-13 11:03:29.496944 |
| Unknown | 26 | 260.2 | 0.14 | No | 67 | 2018-09-13 11:03:47.653924 |
| Yogita | 26 | 261.1 | 0.14 | Yes | 66 | 2018-09-13 11:03:53.520794 |
| Yogita | 26 | 260 | 0.14 | Yes | 66 | 2018-09-13 11:03:58.639831 |

### 3.4 Data Preprocessing and Feature Scaling

First, respective libraries like pandas, numpy, matplotlib, preprocessing, and classifiers for ML are imported with the help of python programming. A dataset that is in the form of Comma Separated Values (CSV) file is imported using python. Dataset is checked for missing values. It is free from missing values. After data preprocessing like handling of missing values, PIR categorical feature is converted into additional numeric features using either dummy variables or one hot encoding concept. Most of the ML algorithms cannot work directly on the categorical data except DT. Before the utilization of any ML algorithm, it is necessary to apply one-hot encoding to convert categorical features into numeric. The PIR feature is converted into two more additional features 'PIR_Yes' and 'PIR_No', which have values either 0 or 1 based on their categories in the original dataset.

Dataset is divided into training and testing samples to apply different feature scaling techniques. Multi-class classification problem consists of 7 features out of which 1 is a class label in the form of 'Person' feature and remaining features like Temp, LDR, Gas, Hum, PIR_Yes, and PIR_No are input features. Different feature scaling techniques are applied to analyze the performance of the ML models. Standard Scalar or Min-Max Scalar is one of the popular and efficient feature scaling techniques, which is applied to enhance the performance of the system. It is less sensitive to outliers, retains the meaningful information of the outliers, and normally distributes the data objects to increase the overall performance of the system.

```
Algorithm 2 Data preprocessing and feature scaling
1.  Start
2.  Import pandas and numpy libraries.
3.  Import the Cloud dataset.
4.  Drop the 'timestamp' feature from the Cloud dataset.
5.  Test the dataset for missing values.
6.  Handle the missing values, if any.
```

```
7.  Convert 'PIR' categorical feature into numeric using one-hot encoding.
8.  Drop the 'PIR' feature from the Cloud dataset.
9.  Extract input features except for the 'Person' feature and
    store in X.
10. Extract the 'Person' feature as an output feature and store in y.
11. Divide the dataset into training and testing
12. Import MinMaxScaler from Sklearn
13. Perform feature scaling using MinMaxScaler
14. End
```

Algorithm 2 gives complete procedural steps to perform data preprocessing and feature scaling techniques. At the beginning pandas and numpy libraries are imported in step 2. The Cloud dataset is imported in step 3. First, the 'Timestamp' feature is removed from the dataset in step 4. Dataset is free from the missing values. 'PIR' Categorical feature is converted into additional features using the one-hot encoding of pandas in step 7 and the 'PIR' feature is removed in step 8. Input and output features are extracted into X and y variables respectively from the final dataset in steps 9 and 10. X is an independent variable and y is a dependent variable. Input and output features that are available in X and y are divided into training and testing datasets using the train_test_split () method of model_selection library of sklearn in step 11. MinMaxScaler feature scaling is applied to input features to bring all sensor values onto the same scale from 0 to 1 in step 13. Fit () method of MinMaxScaler is applied on training dataset, while Transform () method of MinMaxScaler is applied on training as well as testing datasets which lead to proper feature scaling. The fit () method finds the minimum and maximum values from the respective feature whereas the transform () method converts sensor values into new representation before construction of actual supervised ML models to enhance the performance of the models. Similarly, data preprocessing and feature scaling techniques are applied over the local dataset.

Data scaling is not so significant for DT, Random Forest (RF), and Gradient Boosting (GB), it is essential for Naïve Bayes (NB), Logistic Regression (LR), and K-Nearest Neighbor (KNN) to work objective function effectively. Min-max normalization is one of the simplest techniques to bring feature values to standardize in the range from 0 to 1. Han et al. (2012b) presented a formula, which is given in equation (1):

$$x' = \frac{x - \min(x)}{\max(x) - \min(x)} \big(new\_\max(x) - new\_\min(x)\big) + new\_\min(x) \qquad (1)$$

where $x$ is the original value, $x'$ is normalized value, $\min(x)$ is the minimum value and $\max(x)$ is the maximum value of that feature, $new\_\max(x)$ is the new maximum value and $new\_\min(x)$ is the new minimum value.

## 4. DATA ANALYSIS AND RESULT DISCUSSION

The PLMS addresses various problems by providing solutions such as identifying abnormal conditions using threshold values in the programming itself or detecting those conditions using a statistical approach like box-whisker plots and sending alert mail to the user to take corrective action to avoid further loss in the environment. Box-whisker plot gives five number summaries, divides entire data into four parts as well as identifies outliers in the data if any. The first part is from the minimum to first quartile (Q1); the second part is from Q1 to the second quartile (Q2); the third part is from Q2 to the third quartile (Q3), and the fourth part is from Q3 to the maximum. Sensor data values, which are either less than minimum or greater than maximum, are called outliers. Analysis of the box-whisker plot is applied to set threshold values of the sensors to identify abnormal conditions in the environment. The PLMS also plots live sensor values on ThingSpeak open-source Cloud to visualize and understand the physical location. It gives a solution to multi-class classification problems to optimize person prediction using sensor data analysis and determines the most informative sensor. To find the overall performance of the PLMS, the macro average f1-scores are measured. As far as person prediction is concerned, different ML models are compared concerning macro average f1-scores, to select the best predictive model. Time series prediction and analysis problems are also addressed to predict a person using time and date. The PLMS was kept in three different physical locations such as home, academy office, and company office to observe the location in a real-time manner. Three different physical locations are monitored to see environmental effects, if any, to optimize person prediction using sensor data analysis to select the best predictive ML model as well as to optimize person prediction using the timestamp to select the most prominent ML model. The VAR ML model is also constructed to predict sensor values of different time series in a multivariate scenario.

### 4.1 Person Prediction Using Sensor Data Analysis

The Cloud dataset is maintained onto the GoDaddy Cloud, while the local dataset is maintained in the Raspberry-Pi itself. Person prediction data analysis is done for the Cloud dataset as well as the local dataset. The Cloud dataset is small and imbalanced whereas the local dataset is large and more imbalanced. To find the best predictive and robust ML model, three baselines and three advanced ML algorithms' performance is evaluated and compared for Cloud as well as local datasets. Six different supervised ML algorithms like KNN, NB, LR, DT, RF, and GB are applied. Out of six different ML algorithms, the first three models: KNN, NB, and LR are baseline models, whereas DT, RF, and GB are advanced models. Six supervised ML algorithms are summarized as follows:

- KNN

It is a slow learner, which is based on distance measurement techniques as well as the number of nearest neighbors, which is represented by K, from the unknown object. It measures the distance from unknown objects to every known object, finds K nearest neighbors, and a majority vote of the class is assigned to the unknown object. Data preprocessing is required before applying KNN algorithms to bring all feature's values into the same scale to enhance the performance of the model.

- NB

It is based on the Bayes theorem and assumes conditional independence between the features, which calculates the posterior probability of every class and assigns the class with maximum posterior probability to unknown objects. Prior probability and the likelihood of the class play an important role to calculate posterior probability.

- LR

Sigmoid function, which is based on the odds ratio, is used in LR to calculate the probability. If the probability is greater than 0.5, then a positive class is assigned to an unknown object, otherwise, a negative class is assigned.

- DT

Feature selection techniques such as Information Gain, Gain Ratio, and Gini Index are used to construct DT a recursive fashion. Nodes of the DT represent feature names and branches of the node represent outcomes of the features. Leaf node shows the class label of the data object. Tree pruning can be used to avoid the overfitting of the DT model.

- RF

It is an ensemble model, which is based on DT, to avoid overfitting problems of the DT by constructing multiple DT simultaneously. Randomness in the trees is achieved by selecting bootstrap samples from the training dataset and by selecting features randomly at every node of the tree. The majority vote of the class label is assigned to the unknown object.

- GB

It is also an ensemble model and based on DT, which constructs multiple DT sequentially, one after another to learn from the mistakes of the previous DT. It requires proper tuning of the hyper parameters like the number of DT to be constructed, learning rate, and depth of the DT. The cumulative result of the boosting algorithm is considered to assign the class label to unknown objects.

To determine the most informative feature to predict a person, feature importance is also calculated. One vs. all, a variation of multi-class classification, is applied to build the classifiers. One vs. all multi-class classifiers develops five binary classifiers for the Cloud dataset as there are 5 class labels and 6 binary classifiers for the local dataset, which has 6 class labels. The local dataset contains 'No person' as the sixth category of the 'Person' output feature, as irrespective of person recognition sensor reading with 'No person' person label is pushed onto the local server. The performance of the multi-class classification is measured in terms of micro-average f1-scores and macro-average f1-scores. Micro-average precision and recall are calculated from the individual False Positive (FP), True Positive (TP), and False Negative (FN) in n class system. Micro average f1-score is just the harmonic mean of micro average precision and recall in n class system and is useful when there is a need to weigh each sample or forecast evenly.

$$Precision_{micro} = \frac{\sum_{i=1}^{n} TP_i}{\sum_{i=1}^{n} TP_i + FP_i} \quad (2)$$

$$Recall_{micro} = \frac{\sum_{i=1}^{n} TP_i}{\sum_{i=1}^{n} TP_i + FN_i} \quad (3)$$

Macro-average precision and recall are calculated as the average scores of the different n systems. The macro average f1-score is just the harmonic mean of macro average precision and recall in n class system and is useful when there is the necessity to know how the system performs in general across the sets of data.

$$Precision_{macro} = \frac{\sum_{i=1}^{n} Precision_i}{n} \quad (4)$$

$$Recall_{macro} n \frac{\sum_{i=1}^{m} Recall_i}{n} \quad (5)$$

It is necessary to assess the overall performance of multi-class classifiers across all classes using stratified 5-fold cross-validation for Cloud as well as local datasets. All classes are evenly distributed among 5 folds to enhance performance metrics in terms of macro average recall, precision, and f1-score. The harmonic mean of macro average precision and recall is called a macro average f1-score. Algorithm 3 shows DT multi-class classifier to find macro average precision, macro average recall, and macro average f1-score using stratified 5-fold cross-validation for the Cloud dataset. Required arrays are created in step 2. The stratified K-fold module is imported in step 3 from sklearn library. In step 4, an object of StratifiedKFold is created and the number of folds is set to 5, which is required for model construction. An Object of the DT model is created in step 5, and the performance scoring parameter is set to 'precision_macro' in step 6. Macro precisions are calculated in step 7 and stored in the respective array in step 8 for further use to plot the graphs. Similarly, the procedure is repeated with a change in performance scoring parameter to calculate macro average recall and macro average f1-scores and stored in respective arrays to plot the bar chart of the same. The same procedure with the change in model construction for other multiclass classifiers is repeated to find macro average precision, recall, and f1-score of remaining classifiers for the Cloud dataset. A similar strategy is also applied to find macro average precision, recall, and f1-score of all multi-class classifiers for the local dataset.

**Algorithm 3** DT multi-class classifier to find macro average f1-score using stratified 5 fold CV
1. **Start**
2. Create three arrays to store macro average precision, recall, and f1-scores.
3. Import StratifiedKFold module from sklearn library.
4. Create an object of StratifiedKFold with some splits=5 as a parameter.
5. Create an object of the DT classifier.
6. Set scoring parameter as scoring_p = 'precision_macro'
7. Invoke cross_val_score method with objects of DT and StratifiedKFold, input and
   Output features, and scoring parameters.
8. Store the results in a macro precision array.
9. Repeat the procedure to find macro average recall, f1-scores, and store results in arrays.
10. **End**

### 4.1.1 Optimization of Person Prediction Using Sensor Data Analysis for Home Physical Location

The PLMS was also kept in three different physical locations like home, academy, and company office to test the performance of the system. For all three physical locations, the overall performance of the system for complete Cloud and local datasets is measured.

Table 3. Macro average f1-scores of home physical location

|  | Cloud Dataset | | | Local Dataset | | |
| --- | --- | --- | --- | --- | --- | --- |
|  | Macro Average Precision (%) | Macro Average Recall (%) | Macro Average F1-score (%) | Macro Average Precision (%) | Macro Average Recall (%) | Macro Average F1-score (%) |
| NB | 25 | 25 | 20 | 17 | 17 | 17 |
| KNN | 40 | 37 | 39 | NA | NA | NA |
| LR | 15 | 24 | 19 | NA | NA | NA |
| DT | 49 | 50 | **49** | 40 | 38 | **38** |
| RF | 47 | 46 | **47** | 46 | 35 | **38** |
| GB | 38 | 27 | 24 | 16 | 17 | 17 |

The Cloud dataset is the major interest dataset as it contains the name of the recognized person with the 'Person' class label for every data sample. The Cloud dataset is imbalanced and contains five different categories for the 'Person' class label. The local dataset is very large and highly imbalanced as well as most of the data samples of the local dataset; contain No Person value with the 'Person' class label, which leads to unimportant for data analysis to predict a person using sensor data. Although the local dataset is unimportant for data analysis, it is used to check the effect of ML algorithms on a larger dataset. Table 3 shows the macro average f1-scores of the home physical location of different multi-class classifiers for the Cloud as well as local datasets. 'NA' in the table for the local dataset indicates that the classifiers KNN and LR take too much time to find macro average precision, recall, and f1-score of the system. Table 3 shows that macro average f1-scores of DT and RF are 49% and 47%, respectively, and better than any other classifiers for the Cloud dataset. It shows that DT, as well as RF, gives approximately similar and reasonable results for the Cloud dataset. The macro average f1-score is the harmonic mean of macro average precision and recall and gives the overall performance of the system across five categories of the 'Person' class label. Because of imbalanced Cloud dataset and involvement of macro average precision and recall in calculating macro average f1-score, the performance of the DT and RF is reasonably acceptable as compared to other ML algorithms. Table 3 also shows that macro average f1-scores of DT and RF are 38% each and better than any other classifiers for the local dataset. It also shows that macro average f1-scores of DT and RF reveal around similar outcomes for the local dataset.

Figure 2 shows the overall performance of person prediction in terms of macro average precision, recall, and f-scores of home physical location for Cloud and local datasets using stratified fivefold cross-validations. Figure 2(a) shows the performance of six classifiers for the Cloud dataset in which the LR classifier gives the least performance of the PLMS. Macro average precision, recall, and f1-scores of LR multi-class classifiers are 15%, 24%, and 19%, respectively. Similarly, macro average precision, recall, and f1-scores of DT are 49%, 50%, and 49% respectively for the Cloud dataset.

DT and RF classifiers give reasonably better macro average f1-score across all five classes as compared to NB, LR, GB, and KNN classifiers. Macro average f1-scores of DT and RF are 49% and 47%, respectively, which are comparatively greater than other classifiers' macro average f1-scores and leads to the best predictive models for the Cloud dataset.

Figure 2(b) shows the performance of four classifiers apart from LR and KNN as their computational time is elevated for local datasets, i.e., larger datasets. NB and GB classifiers give the least performance of the PLMS whose macro average f1-scores are 17% each.

**Figure 2. Overall performances of classifiers for home physical location**

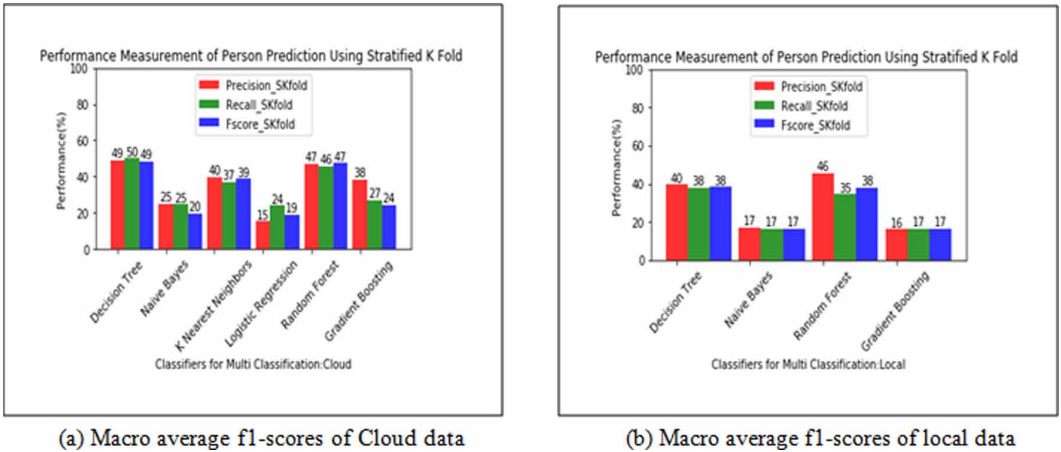

(a) Macro average f1-scores of Cloud data   (b) Macro average f1-scores of local data

Macro average precision, recall, and f1-scores of DT multi-class classifiers are 40%, 38%, and 38%, respectively. Similarly, macro average precision, recall, and f1-scores of RF are 46%, 35%, and 38% respectively for the local dataset.

DT and RF both classifiers give 38% macro average f1-scores, which are greater than the other two models' macro average f1-scores. DT and RF classifiers give remarkable outcomes for small-sized, and less imbalanced Cloud dataset, as well as a large-sized and more imbalanced local dataset to predict a person in the IoT enabled PLMS using sensor data analysis.

### 4.1.2 Optimization of Person Prediction Using Sensor Data Analysis for Academy Office Physical Location

Table 4 shows the macro average f1-scores of the academy office physical location of different multi-class classifiers for the Cloud as well as local datasets. Table 4 shows that macro average f1-scores of DT and RF are 53% and 47%, respectively, and better than any other classifiers for the Cloud dataset. Table 4 also shows that macro average f1-scores of DT and RF are 31% and 29%, respectively, and better than any other classifiers for the local dataset.

**Table 4. Macro average f1-scores of academy office physical location**

|  | Cloud Dataset | | | Local Dataset | | |
|---|---|---|---|---|---|---|
|  | Macro Average Precision (%) | Macro Average Recall (%) | Macro Average F1-score (%) | Macro Average Precision (%) | Macro Average Recall (%) | Macro Average F1-score (%) |
| **NB** | 28 | 43 | 30 | 21 | 38 | 22 |
| **KNN** | 35 | 40 | 37 | NA | NA | NA |
| **LR** | 18 | 25 | 19 | NA | NA | NA |
| **DT** | 54 | 53 | **53** | 33 | 29 | **31** |
| **RF** | 52 | 48 | **47** | 35 | 27 | **29** |
| **GB** | 42 | 40 | 42 | 28 | 26 | 24 |

**Figure 3. Overall performances of classifiers for academy office physical location**

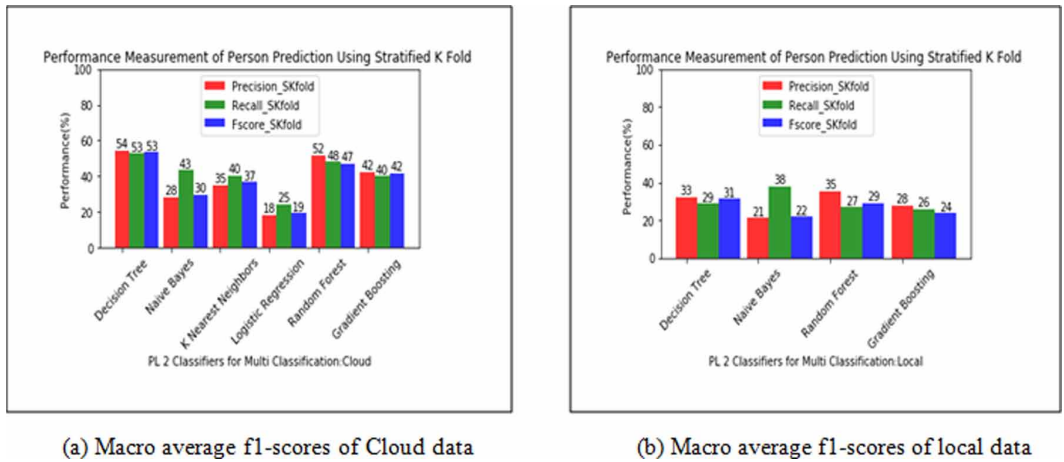

(a) Macro average f1-scores of Cloud data  (b) Macro average f1-scores of local data

Figure 3 shows the overall performance of person prediction in terms of macro average precision, recall, and f-scores of the academy office physical location for Cloud and local datasets using stratified fivefold cross-validations. DT gives remarkable outcomes for Cloud as well as local datasets to predict a person in the IoT enabled PLMS using sensor data analysis.

*4.1.3 Optimization of Person Prediction Using Sensor Data Analysis for Company Office Physical Location*

Table 5 shows the macro average f1-scores of the company office physical location of different multi-class classifiers for Cloud as well as local datasets. Table 5 shows that macro average f1-scores of DT and RF are 56% and 57%, respectively, and better than any other classifiers for the Cloud dataset. Table 5 also shows that macro average f1-scores of DT and RF are 30% and 34%, respectively, and the performance of RF is better than the performance of any other classifiers for the local dataset.

Figure 4 shows the overall performance of person prediction in terms of macro average precision, recall, and f-scores of company office physical location for Cloud and local datasets using stratified fivefold cross-validation. Figure 4(a) shows that macro average f-scores of DT and RF are 56% and

**Table 5. Macro average f1-scores of company office physical location**

|  | **Cloud Dataset** | | | **Local Dataset** | | |
| --- | --- | --- | --- | --- | --- | --- |
|  | **Macro Average Precision (%)** | **Macro Average Recall (%)** | **Macro Average F1-score (%)** | **Macro Average Precision (%)** | **Macro Average Recall (%)** | **Macro Average F1-score (%)** |
| **NB** | 17 | 21 | 08 | 30 | 46 | 30 |
| **KNN** | 53 | 61 | 54 | NA | NA | NA |
| **LR** | 19 | 17 | 09 | NA | NA | NA |
| **DT** | 62 | 54 | **56** | 31 | 32 | **30** |
| **RF** | 57 | 59 | **57** | 35 | 29 | **34** |
| **GB** | 53 | 53 | 45 | 36 | 35 | 33 |

**Figure 4. Overall performances of classifiers for company office physical location**

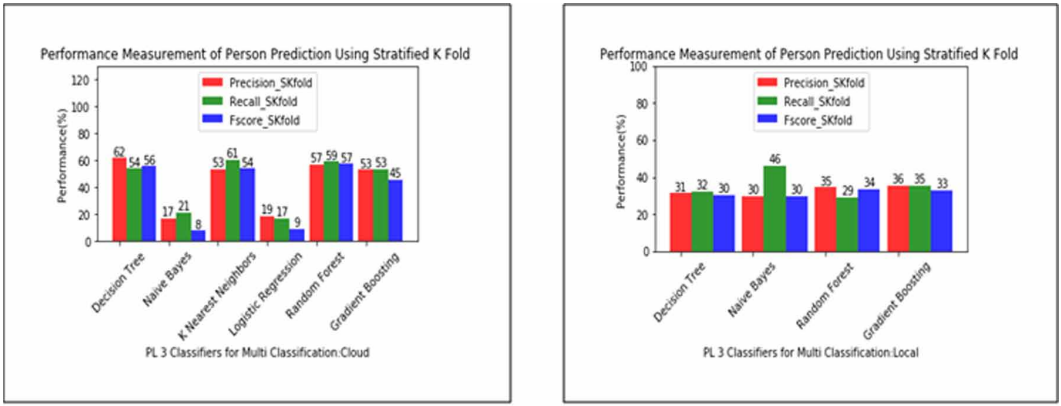

(a) Macro average f1-scores of Cloud data

(b) Macro average f1-scores of local data

57%, respectively, for the Cloud dataset. It shows that RF gives good macro average f-scores for Cloud as well as local datasets.

The Cloud datasets of all three physical locations are also divided into approximately four equal parts separately to check the consistency and reliability of the performance of complete as well as week-wise Cloud datasets. The study also reveals that approximately similar performance is repeated for week-wise Cloud datasets for all three physical locations.

### 4.2 Feature Importance for Person Prediction Using Sensor Data Analysis

DT and RF are summarized with the help of feature importance. Scikit-learn provides important libraries for ML in python. Feature importance is one of the properties of the ML model and associated with the object of that constructed model. Feature importance gives an idea about how one feature is significant from another, and their values differ from 0 to 1 where 0 indicates 'not used at all' and 1 indicates 'completely predicts the target.

Algorithm 4 shows feature importance for the DT classifier. First, the required libraries and the Cloud dataset are imported. After data preprocessing, the dataset is divided into training and testing datasets. An object of the DT classifier is created in step 7. DT classifier is constructed in step 8. Feature array is initialized with the names of all six input features like 'Temp', 'LDR', 'Gas', 'Hum', 'PIR No', and 'PIR Yes' in step 9. The DT model itself gives importance to every feature, which is associated with an object of DT classifier in a variable dt_obj.feature_importances_ and used to plot the graph in step 11. Similarly, feature importance is also calculated for RF classifier.

```
Algorithm 4 Feature Importance for DT classifier
1.  Start
2.  Import pandas and numpy libraries
3.  Import model_selection module from sklearn library.
4.  Import the Cloud dataset
5.  Perform data preprocessing
6.  Split the dataset into training and testing
7.  Create an object of the DT classifier and store it in dt_obj.
8.  Construct the DT classifier model.
9.  Initialize the feature array with the names of six input
    features like    'Temp', 'LDR', 'Gas', 'Hum', 'PIR No', and
    'PIR Yes'.
```

**Figure 5. Person prediction feature importance for home physical location**

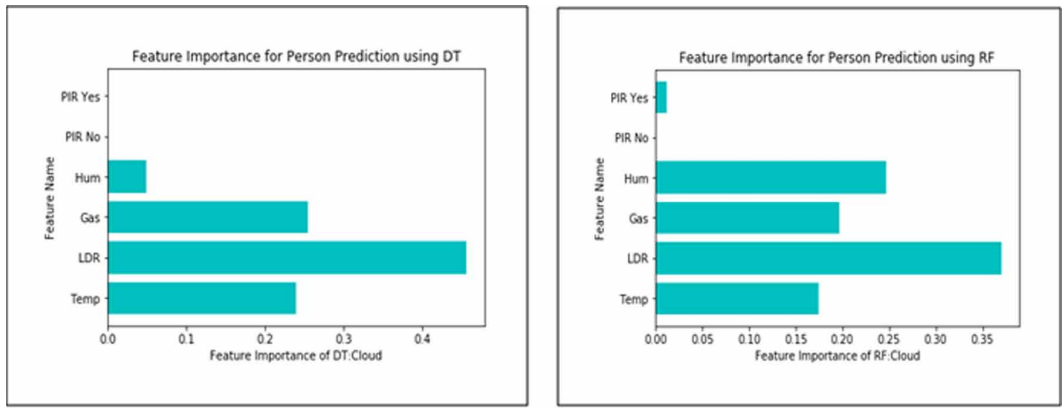

(a) Feature importance of DT    (b) Feature importance of RF

```
10. Set feature importance = dt_obj.feature_importances_
11. Plot horizontal bar using feature_importances_ property of the DT object
12. End
```

### 4.2.1 Person Prediction Feature Importance for Home Physical Location

Figure 5 shows person prediction feature importance of DT and RF for home physical location. Figure 5 depicts that LDR is the most important feature than any other feature in DT as well as RF and confirms our surveillance in analyzing the tree. Figure 5(a) shows that feature importance value for LDR is 0.44 for DT, and Figure 5(b) shows that it is 0.37 for RF. LDR is one of the most important features among all other features to predict a person effectively in the IoT enabled PLMS. LDR feature is followed by a Gas feature in the case of DT whereas it is followed by Humidity feature in case of RF.

### 4.2.2 Person Prediction Feature Importance for Academy Office Physical Location

As discussed earlier, the PLMS was also kept in other physical locations such as the academy and company office to test the consistency and reliability of the system. For all three locations, the most informative feature among various features is found for complete Cloud datasets for DT as well as RF. Experimental results show that most of the time, LDR is the most informative feature to predict a person using sensor data analysis for DT as well as RF for a complete Cloud dataset of three physical locations.

Figure 6 shows person prediction feature importance of DT and RF for the academy office physical location. Figure 6(a) shows that Gas is the most important feature of DT. Figure 6(b) shows that LDR is the most important feature among other features, and the feature importance value of LDR is at least 0.44. As RF is an improvement over the DT, RF gives reliable results than DT.

### 4.2.3 Person Prediction Feature Importance for Company Office Physical Location

Figure 7 shows person prediction feature importance of DT and RF for company office physical location. Figure 7(a) and Figure 7(b) show that the LDR is the most informative feature with values 0.85 and 0.65, respectively, for DT and RF. Results of feature importance of all three physical locations for complete Cloud datasets indicate that the LDR is the most consistent and reliable feature to predict a person using sensor data analysis.

The Cloud datasets of all three physical locations are also divided into approximately four equal parts separately to check the consistency and reliability of the feature importance of complete

**Figure 6. Person prediction feature importance for academy office physical location**

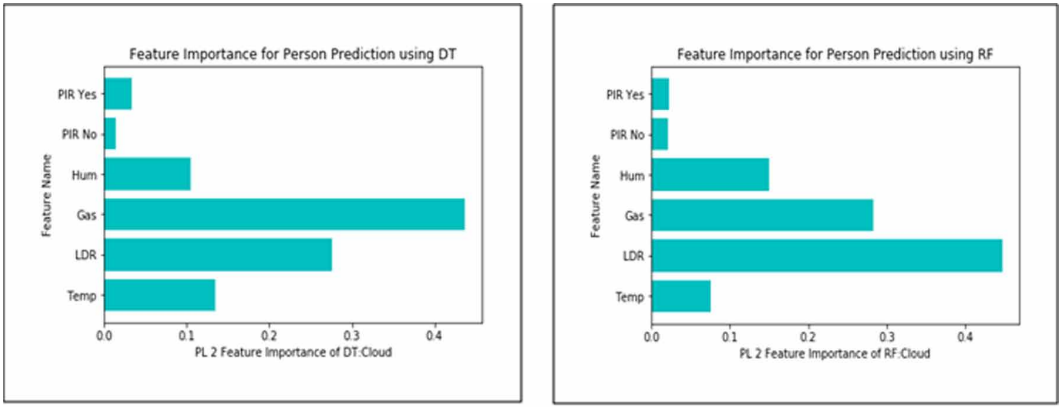

**Figure 7. Person prediction feature importance for company office physical location**

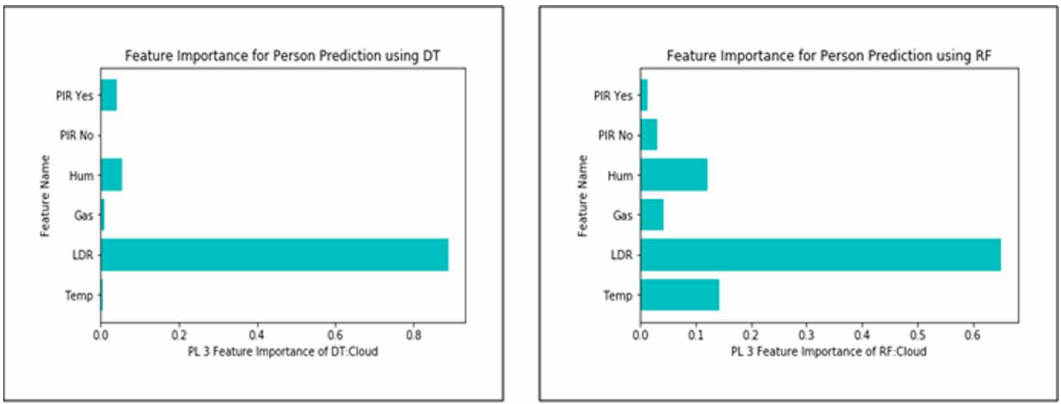

Cloud datasets with week-wise the Cloud datasets. The study also reveals that approximately similar performance is repeated for every week-wise Cloud datasets for all three physical locations.

### 4.3 Optimization of Person Prediction Using Timestamp (Time and Date)

From the real-time collected sensor Cloud dataset, timestamp and person class labels are removed to get only timestamp and class labels, respectively. Timestamp acts as an input feature which is accessible as date and time. The person class label is the output categorical feature. To predict a person using the timestamp as an input feature, it is necessary to do factorization of output feature using pandas. On personal computers, dates are recorded according to the Portable Operating System Interface (POSIX) time understanding, which is the number of seconds, slipped by since January 1970 00:00:00. To apply different supervised ML algorithms, the timestamp is renewed into the number of seconds since the first day of January 1970, and the output categorical person feature is factorized to different labels using pandas. Different ML algorithms like DT, KNN, NB, and RF are applied to see the performance of the PLMS. To select the best predictive model, accuracies of all classifiers are compared to predict a person using a timestamp.

```
Algorithm 5 DT classifier to predict Person using a timestamp
(Date and Time)
1.  Start
2.  Import pandas and numpy libraries.
3.  Import the Cloud dataset.
4.  Import DateTime module.
5.  Extract 'TimeStamp' as an input feature in the Time variable.
6.  Extract 'Person' as an output feature in the Person variable.
7.  Use the factorization method of pandas to assign numeric
    values to categories of Person.
8.  Set epoch to DateTime of 01/01/1970.
9.  Convert Time, which is in string form to DateTime using
    strptime () method of DateTime module and store in temp1.
10. Subtract epoch from temp1 and store in td.
11. Count the total number of seconds elapsed from 01/01/1970
    using td and store in temp2.
12. Append temp2 to the new Time1 array.
13. Reshape Time1 to convert numpy matrix into a vector.
14. Construct DT classifier with Time1 as an input feature and
    Person as an output feature.
15. To predict a person, read the timestamp.
16. Convert Time, which is in string form to DateTime using strptime ().
17. Calculate the total number of seconds elapsed from 01/01/1970
    and reshape it into a vector.
18. Apply the ML model, which is created in step 12 to predict a
    person's name.
19. Display the name of the predicted person.
20. End
```

Algorithm 5 shows procedural steps of the DT classifier to predict a person using a timestamp, i.e. the date and time. First, the Cloud dataset which is available in an excel file is read using the read_excel () method of pandas and stored in the variable. Among various features of the dataset, only 'TimeStamp' and 'Person' features are extracted and stored in the variables like Time and Person in steps 5 and 6, respectively. Time represents input feature, whereas Person represents output feature, which is in categorical form and has five distinct class labels such as 'Ajitkumar', 'Yogita', 'Swaroop', 'Pramila', and 'Unknown'. Using the factorize () method of pandas in step 7, the class label Person is factorized into labels which are in numeric form and assigned numbers from 0 to 4 for different categories of Person for all samples and unique which stores all distinct class labels. The epoch variable is set to 01/01/1970 in step 8. The timestamp of every sample, which is in string form, is converted into DateTime form in step 9. The strptime () method, in step 9, of DateTime module is used to parse the string, which represents timestamp in the specific format given by the second parameter of the same method and parsed string is stored in temp1 variable. The first parameter represents the string that one wants to parse. The temp1 variable shows a timestamp in proper format like '%Y-%m-%d %H: %M: %S. %f'. The epoch is subtracted from the temp1 to get the result in the form of the number of days and '%H: %M: %S. %f' elapsed since 01/01/1970 and stored in the variable td in step 10. At the last variable, td is finally converted into the total number of seconds elapsed since 01/01/1970 and stored in the variable temp2 in step 11. After conversion, all samples' timestamps are stored in the Time1. The reshape () method of numpy array is used to convert numpy matrix, i.e. Time1, into a vector in step 13. DT classifier is constructed on the dataset, which has now Time1 as an input feature and Person as an output feature in step 14. Date, for which we want to predict the person's name, is accepted in step 15 and converted into the number of seconds elapsed since 01/01/1970 as well as reshaped into the vector in steps 16 and 17. DT classifier is used to predict the person's name in step 18, and the name along with the timestamp is displayed on the screen in step 19. Similarly, other supervised ML models are created to predict a person using a timestamp.

Figure 8, Figure 9, and Figure 10 show the accuracies of person prediction using a timestamp of four different supervised ML algorithms for the home, academy office, and company office physical locations respectively. The X-axis shows the names of four ML classifiers, and Y-axis shows accuracies in percentage (%). Figure 8 shows that accuracies of DT, KNN, NB, and RF are 83.99%, 81.64%, 38.01%, and 83.29%, respectively. Figure 9 shows that accuracies of DT, KNN, NB, and RF are 88.92%, 84.73%, 68.86%, and 88.02%, respectively. Figure 10 shows that accuracies of DT, KNN, NB, and RF are 80.97%, 49.01%, 38.64%, and 77.27%, respectively. Among four prescient models, DT gives the most magnificent presentation, and NB gives the most noticeably terrible exhibition of the model for person prediction using date and time for all three different physical locations. Results reveal that irrespective of physical environmental conditions, DT is the most consistent and reliable supervised ML model to predict a person using time and date.

### 4.4 Multivariate Time Series Prediction Using Vector Auto Regression (VAR)

The time series like Temperature (Temp), Humidity (Hum), Light Dependent Resistor (LDR), and Gas are stationary as their statistical properties don't change over a period of time. The study is also performed to predict different sensor values in a multivariate scenario. In VAR, every time series is represented as a function of its own lagged values along with the lagged values of other time series in the system, which gives multiple equations with one equation per time series. In this study, VAR consists of four equations, as the PLMS contains four different time series such as Temp, Hum, LDR, and Gas. The system of equations with four time series is given in equations 6, 7, 8, and 9. For simplicity, the order of the VAR is considered one, as only the first lag values of all time series are represented in all equations.

**Figure 8. Person prediction using a timestamp for home physical location**

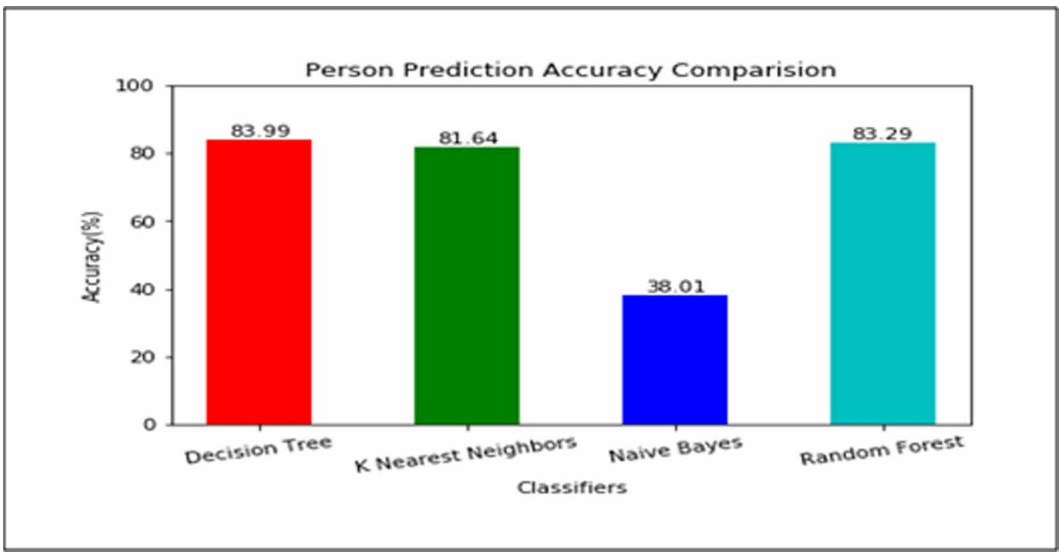

**Figure 9. Person prediction using a timestamp for academy office physical location**

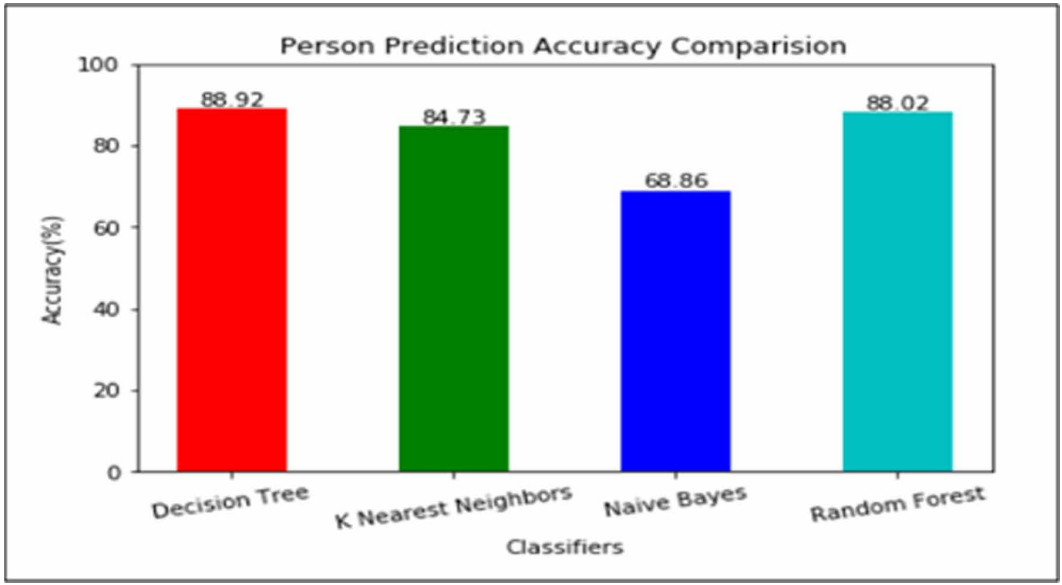

**Figure 10. Person prediction using a timestamp for company office physical location**

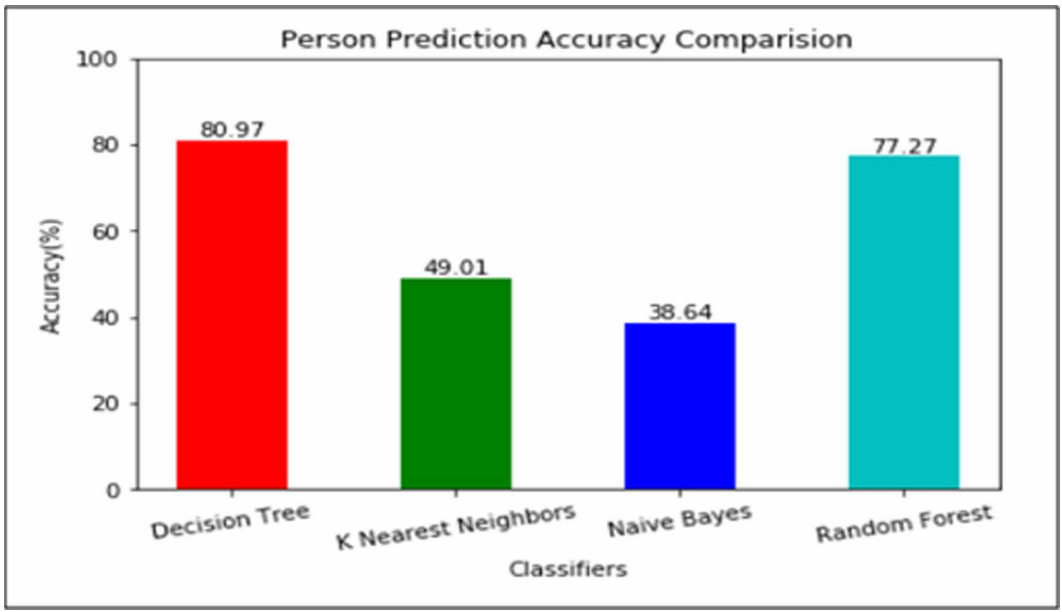

$$Y_{Temp,t=} \propto_1 + \propto_{11} Y_{Temp,t-1} + \propto_{12} Y_{Hum,t-1} + \propto_{13} Y_{LDR,t-1} + \propto_{14} Y_{Gas,t-1} + \in_{1,t} \quad (6)$$

$$Y_{Hum,t=} \propto_2 + \propto_{21} Y_{Temp,t-1} + \propto_{22} Y_{Hum,t-1} + \propto_{23} Y_{LDR,t-1} + \propto_{24} Y_{Gas,t-1} + \in_{2,t} \quad (7)$$

$$Y_{LDR,t=} \propto_3 + \propto_{31} Y_{Temp,t-1} + \propto_{32} Y_{Hum,t-1} + \propto_{33} Y_{LDR,t-1} + \propto_{34} Y_{Gas,t-1} + \in_{3,t} \quad (8)$$

$$Y_{Gas,t=} \propto_4 + \propto_{41} Y_{Temp,t-1} + \propto_{42} Y_{Hum,t-1} + \propto_{43} Y_{LDR,t-1} + \propto_{44} Y_{Gas,t-1} + \in_{4,t} \quad (9)$$

In equation (6), $Y_{Temp,t}$ shows the predicted value of the Temperature sensor at timestamp t. $\propto_1$ is the intercept and $\propto_{11}$, $\propto_{12}$, $\propto_{13}$, and $\propto_{14}$ are the coefficients of the first lags of the four time series. $Y_{Temp,t-1}$ is the first lag value of the Temperature time series; $Y_{Hum,t-1}$ is the first lag value of the Humidity time series; $Y_{LDR,t-1}$ is the first lag value of the LDR time series, and $Y_{Gas,t-1}$ is the first lag value of the Gas time series. $\in_{1,t}$ is the error term. Similarly, these terminologies are used to represent the terms of the remaining equations as well.

**Algorithm 6** Construction of the VAR model to predict the values of the time series
1. **Start**
2. Import the Cloud dataset that contains four time series.
3. Visualize all four time series to observe the trend pattern over the timestamp.
4. Apply Granger's Causation Test to determine the relationship between the four time series.
5. Apply the Augmented Dickey-Fuller (ADF) test to determine whether the time series is stationary or not.
6. Find out the order of the VAR model using the Akaike Information Criterion (AIC) to get the number of lags of the time series.
7. Split the time series into training and testing datasets.
8. Construct the VAR model of order P, which is identified in Step 6, using a training dataset.
9. Use the testing dataset to evaluate the VAR model of order P.
10. Visualize the actual and predicted results.
11. **End**

Algorithm 6 shows the procedural steps of the VAR model that are applied to all four time series of the PLMS. Algorithmic steps of the VAR model show that all four time series influence each other and stationary. VAR model with order three is constructed to predict the time series sensor values. For sake of simplicity, only the results of the home physical location with the last five actual and predicted values are displayed.

**Figure 11. Prediction of temperature time series using VAR**

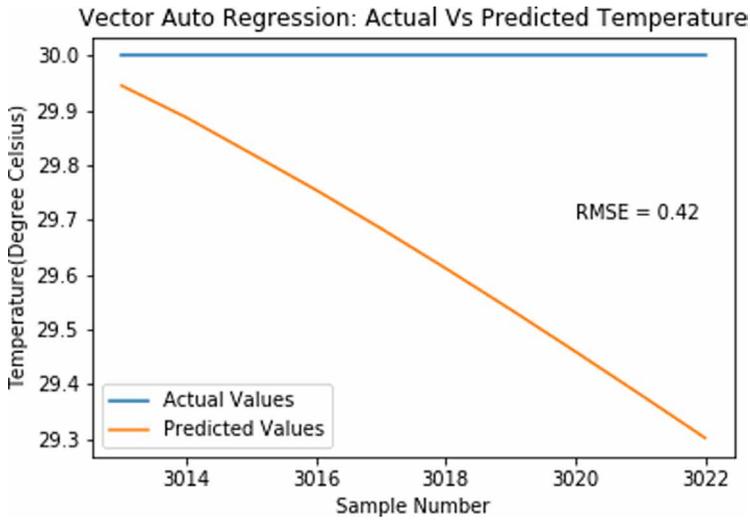

Figure 11 shows the temperature time series prediction using the VAR model. The X-axis represents sample numbers and Y-axis represents Temperature in degree Celsius. Actual and predicted samples are shown using blue and orange colors respectively. RMSE of the temperature series prediction is 0.42.

Figure 12, Figure 13, and Figure 14 show humidity, LDR, and Gas time series prediction using the VAR model respectively. RMSE of the humidity, LDR, and Gas time series are 11.89, 39.99, and 0.01 respectively. RMSE values vary from sensor to sensor because different sensors have their scale of measurements with a variable range of values.

**Figure 12. Prediction of humidity time series using VAR**

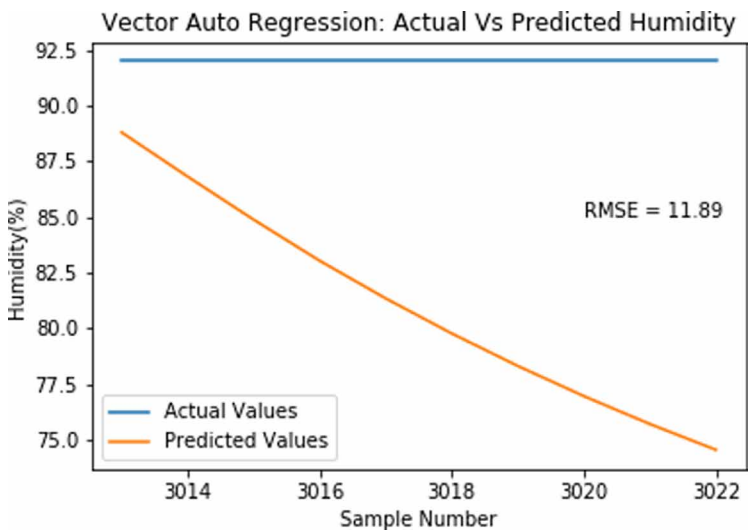

**Figure 13. Prediction of LDR time series using VAR**

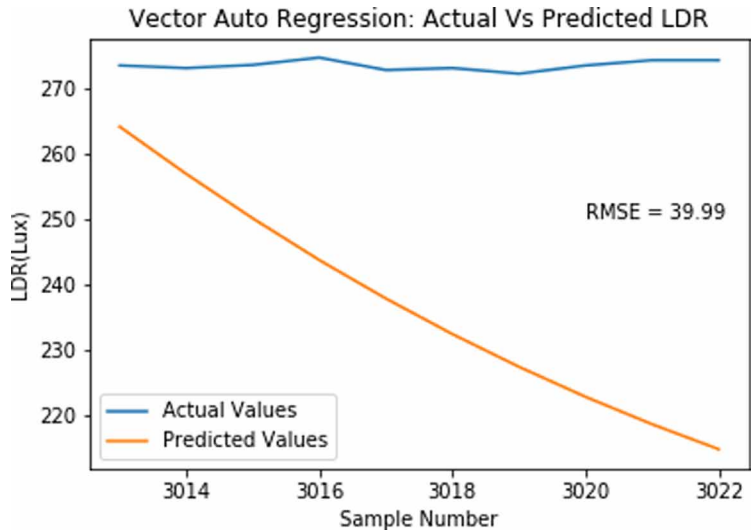

**Figure 14. Prediction of gas time series using VAR**

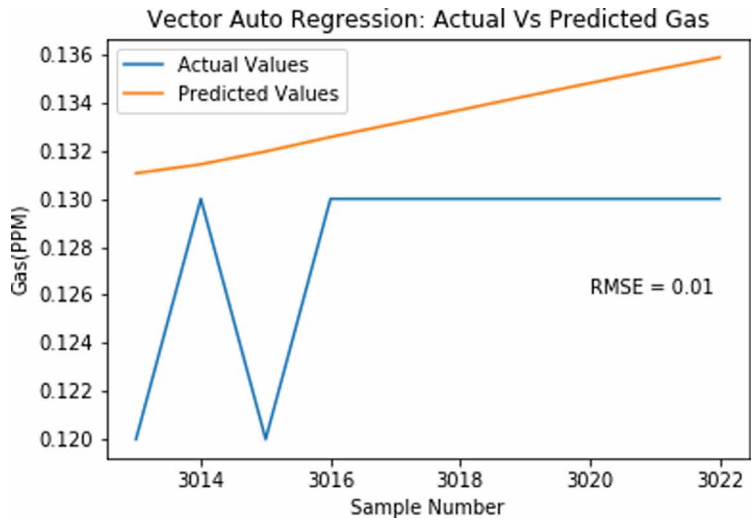

## 5. CONCLUSION AND FUTURE SCOPE

Sensor data analysis with the help of images of the human face for human presence detection and recognition is valuable for affirmation of unusual conditions in the environment. PLMS leads to optimization, as edge computing is performed and only the sensitive information is pushed onto the Cloud to reduce the bandwidth requirement and storage cost onto the Cloud. Experimental results show that for small-sized and imbalanced Cloud dataset as well as large-sized and more imbalanced local datasets, among six multi-class classifiers, DT and RF give a reasonable and approximately similar overall performance in terms of macro average f1-scores to optimize person prediction using sensor data analysis for three different physical locations. Results depict that LDR is the most consistent and reliable informative feature of DT and RF to predict a person using sensor data analysis for complete as well as week-wise Cloud datasets of three different physical locations. The proposed system is novel and prominent to monitor the IoT enabled PLMS to predict a person using a timestamp. The study also reveals that DT is the most consistent and reliable ML model, which gives 83.99%, 88.92%, and 80.97% reasonable accuracies for home, academy office, and company office physical locations respectively, to predict a person using date and time. VAR model with order three gives reasonably good RMSE to predict four time series, i.e., Temperature, Humidity, LDR, and Gas that are stationary and bi-directional in multivariate time series scenario.

Any IoT enabled industrial automation system can use the sensor data analysis techniques used in this work to predict a person's name and sensor values using a timestamp. The system can be useful in monitoring any sensitive biotechnology or life sciences laboratory, military premises where the temperature, humidity, and physical entrance of the restricted person are not allowed.

The PLMS requires a continuous supply of electricity and will not detect a person if the intensity of light is very poor as well as it may mislead face recognition for similar faces. System performance can be extended by incorporating additional sensors like pressure and ultrasonic to get more information. The PLMS can also be placed on the premises for different seasons like summer, rainy, and winter to monitor the location so that a more powerful and robust ML model can be constructed from a huge dataset.

*Ajitkumar S. Shitole (PhD) completed Ph.D. in Computer Science Engineering from Amity University Maharashtra, Mumbai. He has published more than 25 research papers in various International / National Journals / Conferences. His area of interest is Data Mining, Data Science, Machine Learning, and Algorithms. Currently he is working as an Associate Professor in Computer Engineering Department at Hope Foundation's International Institute of Information Technology, Hinjawadi, Pune. He has published two books titled "Design and Analysis of Algorithms" for SPPU and BATU.*

*Manoj Devare (PhD) currently holds position of Associate Professor at Amity Institute of Information Technology, Amity University Mumbai. He has been served as Post Doctorate Fellow at Centre of Excellence on HPC, University of Calabria, Italy. His research deals with Multi-Scale High Performance Computing in Grids, Virtualization, and Clouds. He is also involved in the review process of the International Journals papers, edited chapters from the reputed journals like Elsevier's Future Generation Computer Systems (FGCS), Springer's review system, International Journal of Computing, IEEE Transactions on Parallel and Distributed Systems (TPDS), and SCIT International Journal.*